\documentclass[sigconf, nonacm]{acmart}
\AtBeginDocument{%
  }

\usepackage[ruled, vlined, noend, linesnumbered]{algorithm2e}

\usepackage{enumitem}
\setlist{leftmargin=5.5mm}

\usepackage{subcaption}
\usepackage{multirow}
\usepackage{multicol}

\usepackage[utf8]{inputenc} 
\usepackage[T1]{fontenc}    
\usepackage{hyperref}       
\usepackage{url}            
\usepackage{booktabs}       
\usepackage{amsfonts}       
\usepackage{nicefrac}       
\usepackage{microtype}      
\usepackage{colortbl}
\usepackage[usestackEOL]{stackengine}
\usepackage{makecell}
\usepackage{balance}
\usepackage{lipsum}

\newcommand\blfootnote[1]{%
  \begingroup
  \renewcommand\thefootnote{}\footnote{#1}%
  \addtocounter{footnote}{-1}%
  \endgroup
}

\SetCommentSty{mycommfont}

\usepackage{xcolor}

\makeatletter
\patchcmd{\@algocf@start}
  {-1.5em}
  {0pt}
  {}{}
\makeatother

\usepackage{afterpage}
\usepackage{multicol}

\begin{document}

\title[Graph Feature Preprocessor: Real-time Subgraph-based Feature Extraction for Financial Crime Detection]{Graph Feature Preprocessor: Real-time Subgraph-based\\ Feature Extraction for Financial Crime Detection}

\author[J. Blanu\v{s}a]{Jovan Blanu\v{s}a}
\orcid{0000-0003-4915-6551}
\affiliation{%
  \institution{IBM Research Europe}
  \city{Zurich}
  \country{Switzerland}
}
\email{jov@zurich.ibm.com}

\author[M. Cravero~Baraja]{Maximo~Cravero~Baraja}
\orcid{0009-0001-4580-936X}
\affiliation{%
  \institution{Caltech}
  \city{Pasadena}
  \state{CA}
  \country{USA}
}
\email{mcravero@caltech.edu}

\author[A. Anghel]{Andreea Anghel}
\orcid{0000-0002-6842-9036}
\affiliation{%
  \institution{IBM Research Europe}
  \city{Zurich}
  \country{Switzerland}
}
\email{aan@zurich.ibm.com}

\author[L. von~Niederh{\"a}usern]{Luc~von~Niederh{\"a}usern}
\orcid{0009-0000-0916-5985}
\affiliation{%
  \institution{IBM Research Europe}
  \city{Zurich}
  \country{Switzerland}
}
\email{lvn@zurich.ibm.com}

\author[E. Altman]{Erik Altman}
\orcid{0009-0001-0978-0360}
\affiliation{%
  \institution{IBM Watson Research}
  \city{Yorktown Heights,}
  \state{~NY,}
  \country{~USA}
}
\email{ealtman@us.ibm.com}

\author[H. Pozidis]{Haris Pozidis}
\orcid{0000-0001-5084-6651}
\affiliation{%
  \institution{IBM Research Europe}
  \city{Zurich}
  \country{Switzerland}
}
\email{hap@zurich.ibm.com}

\author[K. Atasu]{Kubilay Atasu}
\orcid{0000-0002-4315-6780}
\affiliation{%
  \institution{TU Delft}
  \city{Delft}
  \country{Netherlands}
}
\email{kubilay.atasu@tudelft.nl}
\begin{abstract}
In this paper, we present \textit{Graph Feature Preprocessor}, a software library for detecting typical money laundering patterns in financial transaction graphs in real time.
These patterns are used to produce a rich set of transaction features for downstream machine learning training and inference tasks such as detection of fraudulent financial transactions.
We show that our enriched transaction features dramatically improve the prediction accuracy of gradient-boosting-based machine learning models.
Our library exploits multicore parallelism, maintains a dynamic in-memory graph, and efficiently mines subgraph patterns in the incoming transaction stream, which enables it to be operated in a streaming manner.
Our solution, which combines our Graph Feature Preprocessor and gradient-boosting-based machine learning models, can detect illicit transactions with higher minority-class F1 scores than standard graph neural networks in anti-money laundering and phishing datasets.
In addition, the end-to-end throughput rate of our solution executed on a multicore CPU outperforms the graph neural network baselines executed on a powerful V100 GPU.
Overall, the combination of high accuracy, a high throughput rate, and low latency of our solution demonstrates the practical value of our library in real-world applications.
To appear as a conference paper at ACM ICAIF'24.
\end{abstract}

\maketitle

\vspace{-.05in}
\section{Introduction}

 \begin{figure}[t!]
    \centering
    \includegraphics[width=0.95\linewidth]{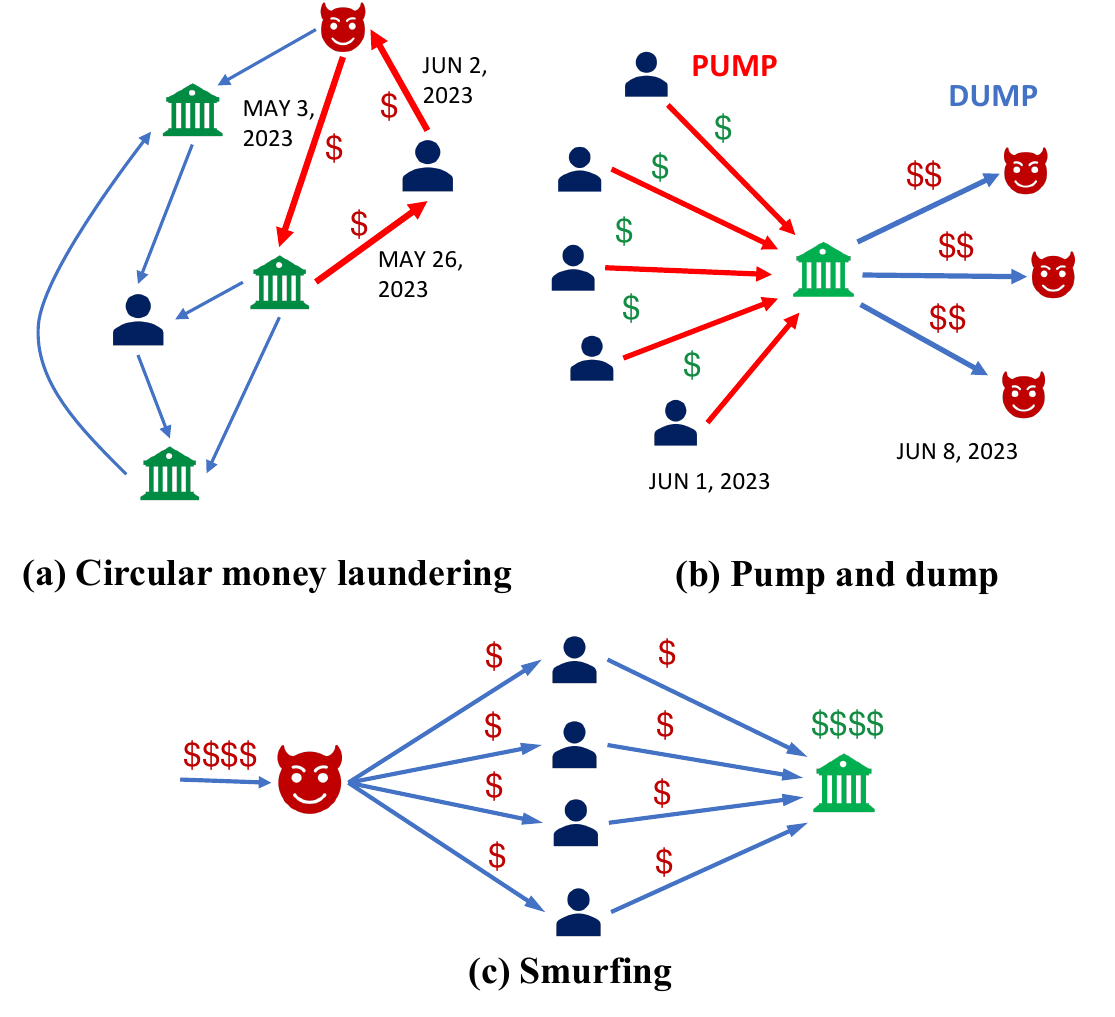}
\vspace{-.15in}
    \caption{Crime patterns in financial transaction graphs.}
    \label{fig:finfraud}
\vspace{-.2in}
\end{figure}

Financial transactions serve as records documenting the movement of financial funds between accounts.
Typically, these transactions are captured in a tabular format, where each row represents a distinct financial transaction, and columns represent basic transaction features such as timestamp, source account, target account, amount transferred, currency, and payment type~\cite{altman_realistic_2023}.
While this tabular representation offers a structured view of the data, a more insightful approach emerges when financial transactions are represented as graphs by treating transactions as edges and accounts as vertices of a graph, as illustrated in Figure~\ref{fig:finfraud}.
Such a graph representation enables analysts to uncover insights that may not be immediately apparent in tabular formats.
As a result, financial transaction graphs facilitate the efficient analysis and interpretation of complex financial data, aiding in the detection of financial crime~\cite{nicholls_financial_2021, corselli_italy_2023}.
\blfootnote{This work was performed when Maximo Cravero Baraja and Kubilay Atasu were with IBM Research Europe, Zurich, Switzerland.}

Subgraph patterns in financial transaction graphs can often serve as indicators of financial crime. 
A \textit{simple cycle}~\cite{mateti_algorithms_1976}, depicted in Figure~\ref{fig:finfraud}a, is one such pattern and represents a sequence of transactions that transfer funds from one bank account back to the same account. 
Such a cycle can be an indicator of financial crimes such as money laundering, tax avoidance~\cite{hajdu_temporal_2020, AMLSim}, credit card frauds~\cite{qiu_real-time_2018, nicholls_financial_2021}, or circular trading used for stock price manipulation~\cite{palshikar_collusion_2008, islam_approach_2009, jiang_trading_2013}. 
In addition, a \textit{gather-scatter} pattern, illustrated in Figure~\ref{fig:finfraud}b, can suggest a \textit{pump and dump} stock manipulation scheme~\cite{nicholls_financial_2021}. 
In this scheme, the stock price of a company is artificially increased through the use of social media to attract other traders for investment.
After the stock price rises sufficiently, malicious traders sell the stocks. 
Due to the artificially inflated stock price, its value drops, and other traders suffer financial losses. 
Furthermore, a \textit{scatter-gather} pattern, depicted in Figure~\ref{fig:finfraud}c, can represent a money laundering tactic called \textit{smurfing}~\cite{kinnison_money_2011, corselli_italy_2023, reuter_chasing_2004, lee_autoaudit_2020,li_flowscope_2020, dong_smurf-based_2021}, in which a malicious actor employs several intermediary accounts (blue nodes in Figure~\ref{fig:finfraud}c) to integrate small sums of illicit funds into the legal banking system. 
Similarly, in cryptocurrency transaction networks, criminals use sophisticated mixing and shuffling schemes to obfuscate the trace of their activities~\cite{liu_knowledge_2021}.
Such schemes can usually be represented by subgraph structures~\cite{ronge_foundations_2021, wang_characteristics_2019, wu_detecting_2021, Whirlpool}. 
The discovery of such suspicious subgraph patterns may enable locating and stopping criminal activities and their perpetrators.

 \begin{figure}[t!]
    \centering
    \includegraphics[width=1.0\linewidth]{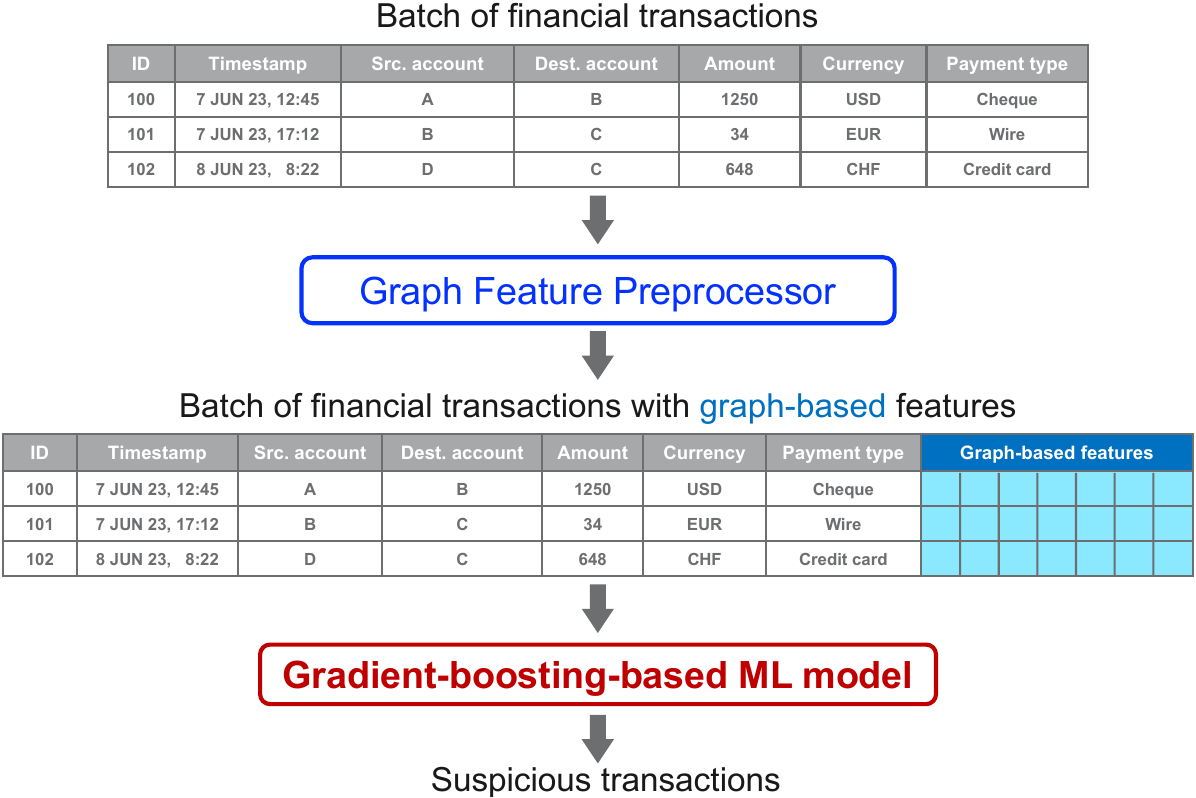}
\vspace{-.1in}
    \caption{The overview of our graph ML pipeline for the detection of suspicious financial transactions.}
    \label{fig:gfp-overview}
\vspace{-.25in}
\end{figure}

Rapid detection and processing of suspicious financial transactions are important to avoid financial losses. 
As financial data is often represented in a tabular format~\cite{altman_realistic_2023}, the fastest and most accurate machine learning models~\cite{grinsztajn_why_2022} for this input format are gradient-boosting-based models~\cite{ke_lightgbm_2017, chen_xgboost_2016}.
However, these models cannot take into account the underlying graph structure and cannot discover graph patterns that could be associated with financial crime. 
Furthermore, a limited set of basic features associated with financial transactions (see Figure~\ref{fig:gfp-overview}) does not provide sufficient information to gradient-boosting-based models for detecting suspicious transactions with sufficient accuracy. 
As a result, the detection of suspicious transactions using these methods~poses~a~challenge.

To overcome the aforementioned limitations, we propose a solution shown in Figure~\ref{fig:gfp-overview}.
Specifically, we develop the \textit{Graph Feature Preprocessor} (GFP) library to produce a rich set of graph-based features for financial transactions.
Our library searches for typical financial crime patterns, such as money laundering cycles and scatter-gather patterns (see Figure~\ref{fig:finfraud}), and encodes these graph patterns into additional columns (i.e., features) of the transaction table.
The transaction table enriched with the graph-based features is then forwarded to a pre-trained gradient-boosting-based machine learning model that performs the classification of financial transactions and detects suspicious transactions.
As a result, the machine learning model is provided with additional transaction features extracted from the financial transaction graph, which facilitates the detection of transactions associated with financial crime.

Our contributions can be summarised as follows:
\begin{itemize}
  \itemsep0em 
  \item We present a graph-based feature extraction library called Graph Feature Preprocessor for enriching the feature set of edges in financial transaction graphs by enumerating suspicious subgraph patterns in graphs as well as by computing various statistical properties of graph vertices.
  We then use this library to develop a graph machine learning (graph ML) pipeline for monitoring financial transaction networks.
  Section~\ref{section:gfp} introduces this~library.

  \item We conduct experiments that demonstrate an improvement of up to $36\%$ in the minority-class F1 score compared to graph neural network (GNN) baselines~\cite{hu2019strategies, corso_principal_2020, cardoso2022laundrograph} for money laundering detection tasks. In addition, we demonstrate that our graph ML pipeline executed using 32 cores of an Intel Xeon processor achieves higher throughput rates compared to those GNN baselines executed on an NVIDIA Tesla V100 GPU. Our experimental evaluation is presented in Section~\ref{sect:results}.

\end{itemize}

The GFP library is publicly available on PyPI as part of Snap ML~\cite{gfp_doc, public_examples, snapml}. In addition, it is offered with IBM\footnote{IBM, the IBM logo, and IBM Cloud Pak are trademarks or registered trademarks of International Business Machines Corporation, in the United States and/or other countries.} mainframe software products \textit{Cloud Pak for Data on Z}~\cite{cp4d} and \textit{AI Toolkit for IBM Z and LinuxONE}~\cite{aitk}. Furthermore, an \textit{AI on IBM Z Anti-Money Laundering Solution Template}~\cite{ai_sol_temp}, which demonstrates how to develop and deploy a graph ML pipeline with GFP using an IBM Z environment, is open-sourced and publicly available\footnote{\url{https://github.com/ambitus/aionz-st-anti-money-laundering}}.

\section{Graph Feature Preprocessor}
\label{section:gfp}

 \begin{figure}[t]
    \centering
	\centerline{
        \includegraphics[width=0.95\columnwidth]{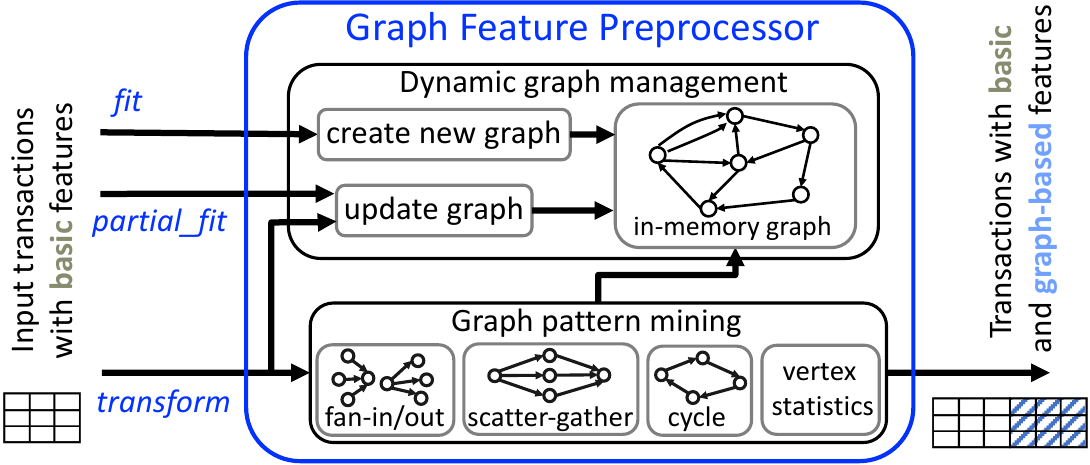}
    }
    \vspace{-.1in}
    \caption{Our Graph Feature Preprocessor is offered as a scikit-learn preprocessor with \emph{fit} and \emph{transform} methods. 
    }
    \vspace{-.15in}
    \label{fig:graph_preprocessor}
\end{figure}

An overview of the Graph Feature Preprocessor (GFP) is given in Figure~\ref{fig:graph_preprocessor}.
It operates in a streaming fashion, receiving as input a batch of transactions with only basic features, such as in Figure~\ref{fig:gfp-overview}, and producing additional graph-based features as output.
GFP stores past financial transactions in an in-memory graph, which is dynamically updated as new transactions are received.
The graph-based features are computed by enumerating subgraph patterns in the graph and by generating statistical properties of the accounts stored in that graph.
GFP can compute the graph-based features across several CPU cores in parallel, which, together with the dynamic graph representation, enables~real-time~feature~extraction.

We have implemented GFP as a scikit-learn preprocessor with the \emph{fit/transform} interface~\cite{sklearn_preprocessor} and made it publicly available on PyPI as part of the Snap ML package~\cite{gfp_doc, snapml, public_examples}.
The main functionality of GFP is implemented by the \emph{transform} function, which is illustrated in Figure~\ref{fig:graph_preprocessor}.
This function inserts a batch of input transactions into the in-memory graph and computes graph-based features for these transactions.
Creating the initial in-memory graph is performed by providing some past transactions as an input to the \emph{fit} function.
The existing in-memory graph can be updated without computing any graph features by using the \emph{partial\_fit} function.
Other standard preprocessor functions supported by GFP are described in the publicly available documentation~\cite{gfp_doc}.
In the rest of this section, we describe the dynamic graph management and graph pattern mining components of GFP (see Figure~\ref{fig:graph_preprocessor}), and we describe how the graph-based features produced by the library~are~encoded.

\subsection{Dynamic Graph Management}
\label{sect:graph_data_struct}

The dynamic graph management component in GFP uses an in-memory graph to represent the financial transaction network.
In this scenario, each account is treated as a graph vertex, and each transaction represents an edge from its source account to its destination account.
As financial transactions typically include a \textit{timestamp} indicating when a transaction was created (see Figure~\ref{fig:gfp-overview}), financial transaction graphs are considered \textit{temporal graphs}~\cite{holme_temporal_2012}.
Furthermore, financial transaction graphs are also \textit{multigraphs}~\cite{Balakrishnan1997}, as there can be several \textit{parallel edges}, i.e.,  edges that connect the same pair of source and destination vertices. 
Hence, our in-memory graph must be capable of representing temporal multigraphs.

To enable the seamless processing of transactions in a streaming fashion, our in-memory graph must support the insertion of new transactions and the removal of outdated transactions. 
We define new transactions as those with timestamps greater than the timestamp of any transaction currently in the in-memory graph.
Outdated transactions are identified as those with timestamps smaller than a value $t_{\mathit{now}} - \delta$, where $t_{\mathit{now}}$ represents the largest timestamp among the transactions in the in-memory graph and $\delta$ denotes a user-defined time window.
Consequently, the in-memory graph retains only transactions that fall within the time window $\left[t_{\mathit{now}}-\delta: t_{\mathit{now}}\right]$, effectively constraining its memory usage.

 \begin{figure}[t!]
    \centering
    \includegraphics[width=1.0\columnwidth]{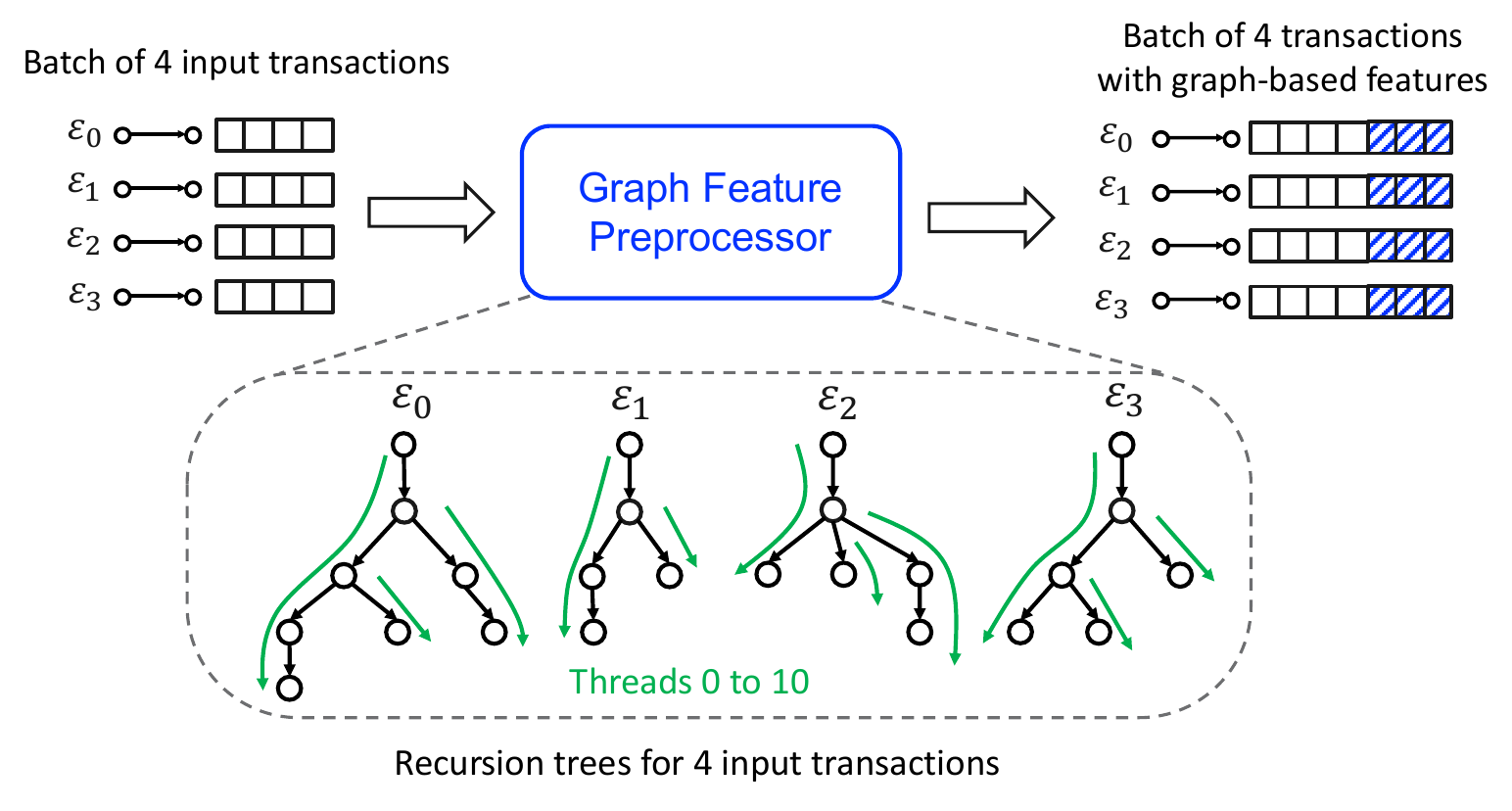}
\vspace{-.1in}
    \caption{Fine-grained parallelism exploited by GFP. The library searches for cycles independently for each input transaction by recursively exploring the transaction graph. The coarse-grained approach would use only four threads, while the fine-grained approach uses eleven threads.}
    \label{fig:gfp-processing}
\vspace{-.1in}
\end{figure}

Our in-memory graph comprises two main data structures: a \textit{transaction log} and an \textit{index}. 
The transaction log, implemented as a double-ended queue, maintains a list of edges sorted in ascending order of their timestamps. 
This data structure facilitates the detection and removal of outdated edges by supporting an $O(1)$ operation for removing the edge with the smallest timestamp.
The index data structure employs an \textit{adjacency list} representation to enable fast access to the neighbours of a vertex~\cite{clrs_2009}. 
Implemented as a vector of hash maps~\cite{stl_umap}, each entry in the vector represents a vertex $v$, and the hash map associated with that vertex $v$ signifies the adjacency list of $v$. 
Vertices are internally mapped to integers in the range of $0, 1, \ldots, n-1$, where $n$ is the number of vertices in the graph.
These integers are used to access the adjacency list of a vertex $v$ in this vector.
Furthermore, each edge can be accessed in $O(1)$ time using the index, facilitating traversal through the graph, as required by the graph pattern mining component.

To support the maintenance of parallel edges in the index, each entry in an adjacency list of the vertex $v$, representing a neighbour $u$ of the vertex $v$, also contains a list of edges connecting $v$ with $u$, referred to as the \textit{parallel edge list}. 
The edges in this list, also implemented as a double-ended queue, are represented with their ID and timestamp, sorted in ascending order of their timestamps. 
For this reason, the operations of inserting new edges and removing the outdated edges can be performed in~$O(1)$~time.

\subsection{Graph Pattern Mining}
\label{sect:gpm}

\begin{algorithm}[t]
\SetAlgoLined
\SetKwProg{Fn}{Function}{}{}
\SetKwProg{Proc}{Procedure}{}{}
\SetKwInOut{InOut}{InOut}
\SetKwComment{Comment}{$\triangleright$\ }{}

\KwIn{$\mathcal{G}$ - the input graph with vertices $\mathcal{V}$ and edges $\mathcal{E}$\DontPrintSemicolon\;
    \hspace*{9.5mm}$\mathit{batch}$ - a batch of edges; 
    \hspace*{1mm}$\delta_p$ - the time window}

 \textcolor{orange}{\textbf{parallel}} \ForEach{$\left( \mathrm{u} \rightarrow \mathrm{v}, t_{uv}\right)$ : $\mathit{batch}$}{  
 \label{algline:sg_oloop}
    $\mathrm{TW} = \left[t_{uv} - \delta_p : t_{uv} \right]$\Comment*[r]{Time window of size $\delta_p$}
    \tcp{The first phase}
    $ N^{+}_{u} = \{ \,\forall x \, | \, \left(u \rightarrow x, \, t_{s}\right)  \in \mathcal{E} \, \land \, t_{s} \in \mathrm{TW} \, \}$\; \label{algline:sg_v_start}
    $ N^{+}_{v} = \{ \,\forall x \, | \, \left(v \rightarrow x, \, t_{s}\right) \in \mathcal{E} \, \land\,  t_{s} \in \mathrm{TW} \, \}$\;
    
    \textcolor{orange}{\textbf{parallel}} \ForEach{$\mathrm{w} : N^{+}_{v}$}{ \label{algline:sg_iloop1}
        $ N^{-}_{w} = \{ \,\forall x \, | \, \left(x \rightarrow w, \, t_{s}\right) \in \mathcal{E} \, \land\,  t_{s} \in \mathrm{TW} \, \}$\;
        $I = N^{+}_{u} \cap N^{-}_{w}$\;
        \lIf{$|I| \geq 2$}{report scatter-gather pattern $\{u, I, w\}$} \label{algline:sg_v_end}
    }
    \tcp{The second phase}
    $ N^{-}_{u} = \{ \,\forall x \, | \, \left(x \rightarrow u, \, t_{s}\right)  \in \mathcal{E} \, \land \, t_{s} \in \mathrm{TW} \, \}$\; \label{algline:sg_u_start}
    $ N^{-}_{v} = \{ \,\forall x \, | \, \left(x \rightarrow v, \, t_{s}\right) \in \mathcal{E} \, \land\,  t_{s} \in \mathrm{TW} \, \}$\;
    
    \textcolor{orange}{\textbf{parallel}} \ForEach{$\mathrm{w} : N^{-}_{u}$}{ \label{algline:sg_iloop2}
        $ N^{+}_{w} = \{ \,\forall x \, | \, \left(w \rightarrow x, \, t_{s}\right) \in \mathcal{E} \, \land\,  t_{s} \in \mathrm{TW} \, \}$\;
        $I = N^{-}_{v} \cap N^{+}_{w}$\;
        \lIf{$|I| \geq 2$}{report scatter-gather pattern $\{w, I, v\}$} \label{algline:sg_u_end}
    }
}
\caption{ScatterGatherStream $\left(\mathcal{G}(\mathcal{V}, \mathcal{E}), \mathit{batch}, \delta_p\right)$}
\label{algo:scat-gat}
\end{algorithm}

The task of the graph pattern mining component is to produce graph-based features for edges forwarded to the library through the \textit{transform} function.
Two types of graph-based features are supported: \textit{i)} graph-pattern-based features and \textit{ii)} vertex-statistics-based features.

\textbf{Graph-pattern-based features} are computed by extracting graph patterns from the in-memory graph that contain one of the forwarded edges.
Our library extracts the following graph patterns: fan-in, fan-out, scatter-gather, gather-scatter, simple cycle, and temporal cycle.
Fan-in and fan-out patterns refer to patterns defined by a vertex $v$ and all of its incoming and outgoing edges, respectively.
A \textit{gather-scatter} pattern combines a fan-in pattern of the vertex $v$ with a fan-out pattern of the same vertex $v$, as illustrated in Figure~\ref{fig:finfraud}b~\cite{dong_smurf-based_2021}.
A fan-out pattern of a vertex $v$ and a fan-in pattern of a vertex $u$ form a \textit{scatter-gather} pattern, depicted in Figure~\ref{fig:finfraud}c, if the fan-out and the fan-in patterns connect vertices $v$ and $u$, respectively, to the same set of intermediate vertices~\cite{dong_smurf-based_2021} (blue vertices in Figure~\ref{fig:finfraud}c).
A simple cycle is a path from vertex $v$ to the same vertex $v$ without repeated vertices except for the first and last vertex.
Finally, a temporal cycle is a simple cycle with edges ordered in time.

To compute graph-pattern-based features in a streaming manner, our library enumerates new patterns that are formed after inserting the input batch of edges into the graph.
The fan-in and fan-out pattern features of a vertex $v$ that belongs to the input batch are determined by counting the number of outgoing and incoming vertices of $v$, respectively.
These features can be determined in $O(1)$ time by simply querying the size of the hash maps that are implementing the adjacency lists of the vertex $v$ in our index data structure (see Section~\ref{sect:graph_data_struct}).
A gather-scatter pattern is detected implicitly if the fan-in and fan-out of a vertex $v$ are at least two.
Due to space constraints, we omit the description of our algorithm for finding scatter-gather patterns in a streaming manner.

 \begin{figure}[t]
	\centerline{
        \includegraphics[width=0.97\columnwidth]{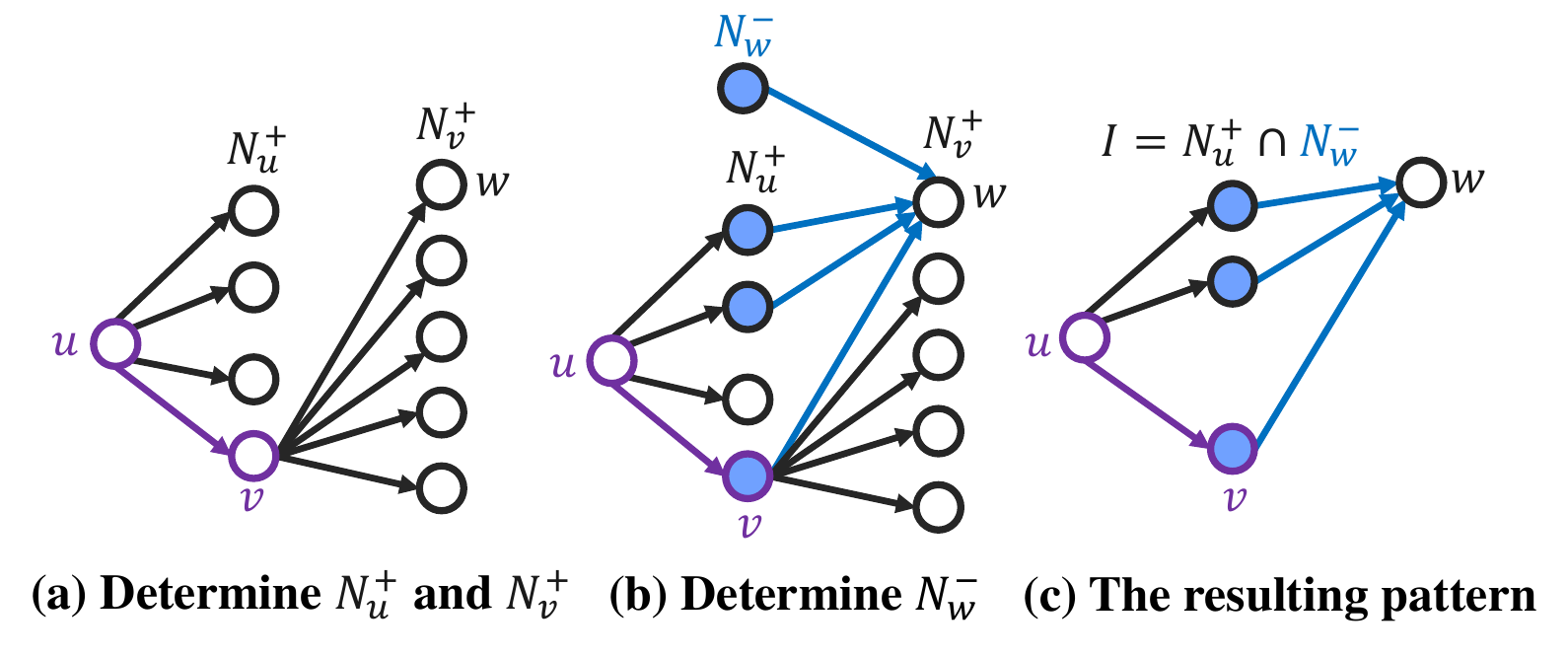}
    }
 	\vspace{-.15in}
    \caption{Enumeration of scatter-gather patterns that contain the edge $u \rightarrow v$ with $v$ being an intermediate vertex.
    }
    \label{fig:findScatGat}
 	\vspace{-.2in}
\end{figure}

To enumerate simple cycles and temporal cycles in a streaming manner, we use fine-grained parallel algorithms introduced in Blanu\v{s}a et al.~\cite{blanusa_scalable_2022, blanusa_cycles_2023}.
These algorithms enable the search for cycles that start from a single edge or a small batch of edges in parallel using several threads.
The benefit of these algorithms is that they can process transactions in small batches with high throughput.
For instance, if the computation of cycles is parallelised by adopting the \textit{coarse-grained} parallel approach, recursive cycle search for each edge of a batch is performed by a different thread.
However, as shown in Blanu\v{s}a et al~\cite{blanusa_cycles_2023, blanusa_scalable_2022} using the coarse-grained approach might result in a suboptimal solution due to the potential workload imbalance across threads.
In contrast, \textit{fine-grained} enumeration algorithms are able to execute the recursive cycle search from a single edge using several threads, as illustrated in Figure~\ref{fig:gfp-processing}, thereby increasing the parallelism.
As a result, even if the input batch contains one transaction, our library would be able to parallelise the search for cycles.

To compute scatter-gather pattern in a streaming manner, we use our algorithm illustrated in Figure~\ref{fig:findScatGat} and presented in Algorithm~\ref{algo:scat-gat}.
In this algorithm, $\left( \mathrm{u} \rightarrow \mathrm{v}, t_{uv}\right)$ denotes a temporal edge with source vertex $u$, target vertex $v$ and timestamp $t_{uv}$.
This algorithm processes each edge $\left( \mathrm{u} \rightarrow \mathrm{v}, t_{uv}\right)$ in the input batch by searching for all scatter-gather patterns that include that edge.
The first and second phase of this algorithm search for scatter-gather patterns that contain $v$ and $u$ as an intermediate vertex, respectively.
In the first phase, we first determine the outgoing neighbours of $u$ and $v$, denoted as $N^{+}_{u}$ and $N^{+}_{v}$, respectively, as shown in Figure~\ref{fig:findScatGat}a.
Then, for each outgoing neighbour $w$ of $v$, we search for incoming neighbours $N^{-}_{w}$ of the vertex $w$, which are represented as filled circles in Figure~\ref{fig:findScatGat}b.
Afterwards, we perform a set intersection between $N^{+}_{u}$ and $N^{-}_{w}$ to find the intermediate vertices $I$ of a scatter gather pattern.
Finally, the algorithm reports the resulting scatter-gather pattern defined with vertices $u$, $w$, and $I$, as shown in Figure~\ref{fig:findScatGat}c.
The second phase of this algorithm, presented in lines~\ref{algline:sg_u_start}--\ref{algline:sg_u_end} of Algorithm~\ref{algo:scat-gat}, is analogous to the first phase, and we omit its description for brevity.
Note that this algorithm can be parallelised in a fine-grained manner by parallelising its loops, as shown in Algorithm~\ref{algo:scat-gat}.

Apart from parallelisation, another method to reduce the time required to find graph patterns is to impose time-window constraints.
In this case, a time window parameter $\delta_p$ can be specified for each graph pattern, in which case the library searches only for patterns whose edges have timestamps greater than or equal to $t_{\mathit{now}} - \delta_p$, where $t_{\mathit{now}}$ represents the largest timestamp among the edges in the in-memory graph.
Additionally, the search for simple cycles can be constrained by limiting their maximal length.

 \begin{figure}[t]
	\centerline{
        \includegraphics[width=0.93\columnwidth]{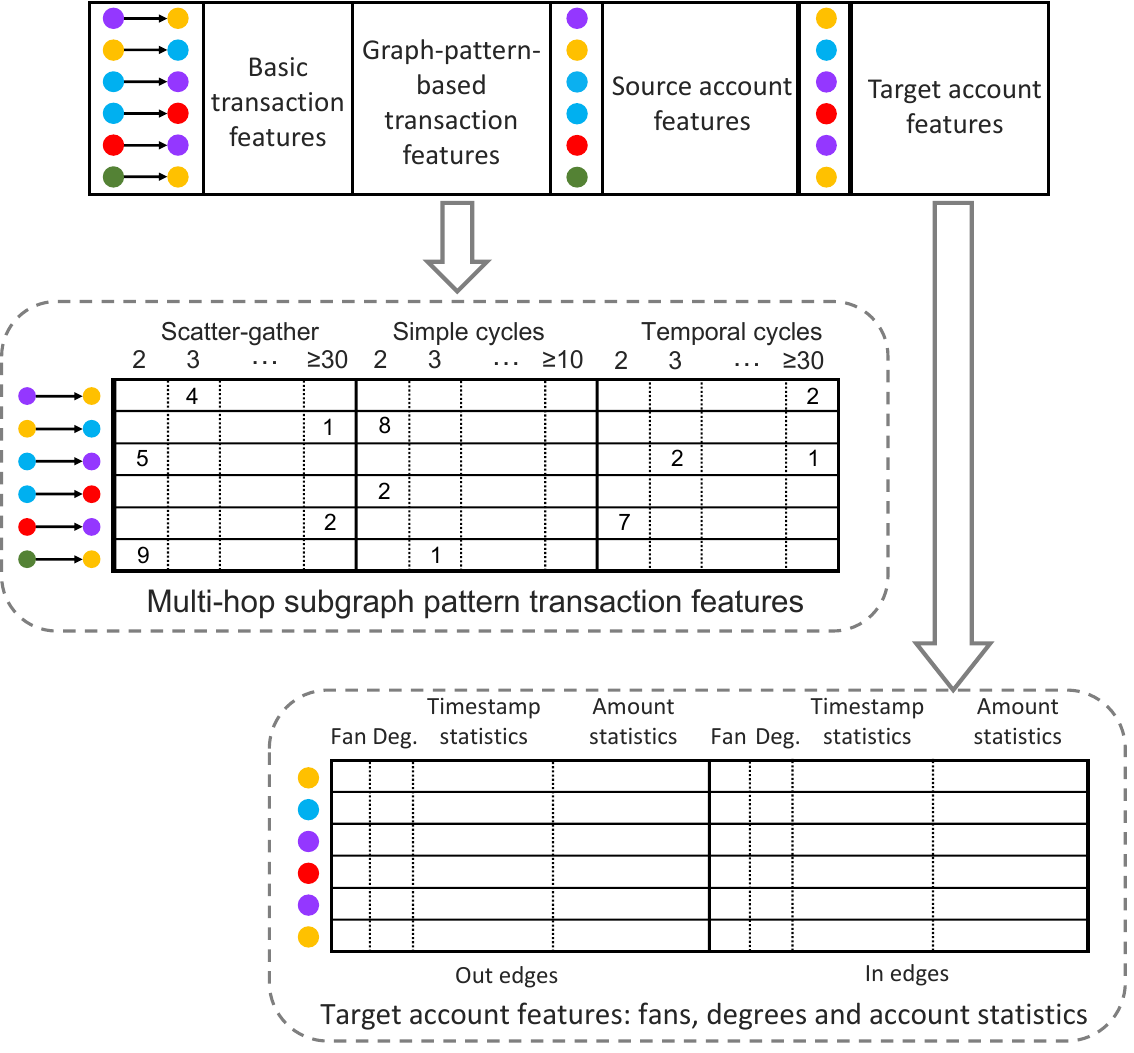}
    }
 	\vspace{-.1in}
    \caption{Feature encoding: scatter-gather patterns are binned according to the number of intermediate vertices they have, and cycles are binned according to their length.
    }
    \label{fig:feature_encoding}
\vspace{-.25in}
\end{figure}

\textbf{Vertex-statistics-based features} are computed for the vertices that appear in the input batch of edges.
For each such vertex $v$, some predefined statistical property can be computed using a  selected basic feature associated with the outgoing edges of $v$ and its incoming edges.
The statistical properties currently supported by our library are: sum, mean, minimum, maximum, median, variance, skew, and kurtosis~\cite{kokoska_crc_2000}. 
For instance, if "Amount" is the selected basic feature used for the calculation of statistical properties, the statistical features include the average and total amount of money an account received or sent.
Combining different statistical feature types with different user-specified basic features in this way extends the feature space~significantly.

Vertex-statistics-based features can be determined in a streaming manner through incremental computation.
For this purpose, our library maintains second, third, and fourth central moments for each vertex of the graph and for each basic feature used for calculating account statistics (e.g., "Amount").
After inserting or removing an edge $u \rightarrow v$, all central moments for $u$ and $v$ are updated incrementally~\cite{finch_incremental_2009, tschumitschew_incremental_2012}.
These central moments are then used to compute the following statistical features: sum, mean, variance, skew, and kurtosis~\cite{kokoska_crc_2000}.
Note that the computation of each aforementioned statistical feature can be performed in $O(1)$ time.
Other statistical features, i.e., minimum, maximum, and median, are simply computed by iterating through the incident edges of a vertex, which is executed in $O(\Delta)$ time per statistical feature, where $\Delta$ is the maximum degree of a vertex in the graph.

\subsection{Feature Encoding}

The encoding of the features produced by the \textit{transform} function of GFP is shown in Figure~\ref{fig:feature_encoding}.
Each row of the output feature table stores the feature vector of a single transaction.
Across different columns of a feature vector, there are basic transaction features, graph-pattern-based transaction features, and the account features of the source and the destination account of the transaction.
The account features consist of vertex-statistics-based features and features based on fan-in and fan-out patterns, both of which are single-hop patterns.
Features based on fan-in and fan-out patterns are computed for each account $v$ and represent the number of accounts connected to $v$ in those patterns.
Graph-pattern-based transaction features are computed using multi-hop subgraph patterns: scatter-gather, hop-constrained simple cycles, and temporal cycles.
For each transaction, our library reports the number of multi-hop subgraph patterns of different sizes that this transaction is part of.
Example features based on multi-hop subgraph patterns are given in Figure~\ref{fig:feature_encoding}, where the first transaction participates in $4$ scatter-gather patterns with $3$ intermediate vertices and in $2$ temporal cycles with $30$ or more edges.
Even though these multi-hop subgraph patterns can also be used to compute account features, computing them as transaction features provides more compact feature vectors.

 \begin{figure*}[ht!]
    \centering
    \includegraphics[width=1.0\linewidth]{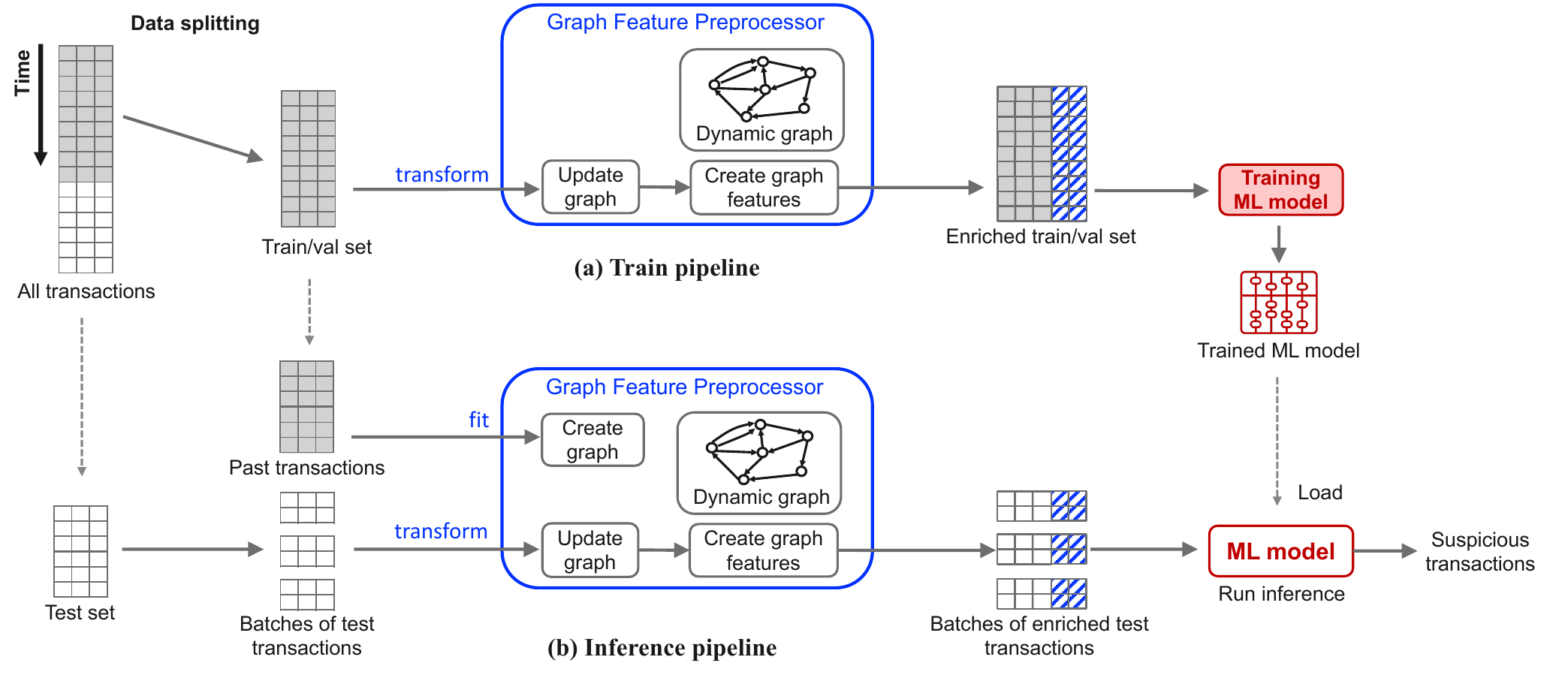}
\vspace{-.15in}
    \caption{Train and inference components of our graph ML pipeline for the detection of suspicious transactions.}
    \label{fig:graphml_pipeline}
\vspace{-.15in}
\end{figure*}

\section{Experimental setup}
\label{subsec:exp-setup}

\begin{table}[t]
\centering
\caption{Datasets used in the experiments.}
\vspace{-.15in}
\addtolength{\tabcolsep}{-1pt}
\begin{tabular}{l|cccc}
\bottomrule
Dataset   & \# nodes      & \# edges      & illicit rate  & time span\\ \hline
AML HI Small         & 0.5 M          & 5 M            & 0.102\%     & 10 days  \\
AML HI Medium        & 2.1 M          & 32 M           & 0.110\%     & 16 days  \\
AML HI Large        &  2.1 M         & 180 M           & 0.124\%     & 97 days  \\ \hline
AML LI Small         & 0.7 M          & 7 M            & 0.051\%     & 10 days  \\
AML LI Medium        & 2.1 M          & 32 M           & 0.051\%     & 16 days  \\
AML LI Large        &  2.1 M         & 180 M           & 0.057\%     & 97 days  \\ \hline
ETH Phishing    & 2.9 M          & 13 M           & 0.278\%      & 1261 days \\ 
\toprule
\end{tabular}
\vspace{-.2in}
\label{tab:datasets}
\end{table}

\textbf{Datasets.} 
Table~\ref{tab:datasets} presents the datasets used in the evaluation.
The AML datasets are publicly available synthetic AML datasets produced by the \textit{AMLworld} generator\cite{altman_realistic_2023}.
These datasets contain transactions labelled as licit or illicit, and, thus, they can be directly used with our graph ML pipeline that performs transaction classification.
The datasets are available in two variants: one with a higher illicit rate (AML HI) and one with a lower illicit rate (AML LI).
In addition, we use the ETH Phishing dataset, which is a real-world Ethereum dataset~\cite{xblockEthereum, ETH_Kaggle} with $1,165$ accounts labelled as phishing.
To enable transaction classification using the ETH Phishing dataset, we label a transaction of this dataset as phishing if its destination account is labelled as phishing.
As a result, $0.278\%$ of Ethereum transactions are labelled as phishing.

\textbf{Baselines.}
We use LightGBM (version 3.1.1)~\cite{ke_lightgbm_2017} and XGBoost (version 1.7.5)~\cite{chen_xgboost_2016} boosting machines, which are widely-used ML models for tabular data, as machine learning models for our graph ML pipeline.
We compare our graph ML pipeline with LightGBM and XGBoost models trained exclusively using basic features, without incorporating features generated by our Graph Feature Preprocessor.
To perform hyper-parameter tuning of these models, we employ a successive halving model tuning approach~\cite{jamieson_best-arm_2014}.
As additional baselines, we use the following graph neural networks (GNNs): Graph Isomorphism Network (GIN)~\cite{xu2018powerful, hu2019strategies}, GIN with edge updates (GIN+EU)~\cite{battaglia2018relational_edgeupdates, cardoso2022laundrograph}, and Principal Neighbourhood Aggregation (PNA)~\cite{corso_principal_2020, velickovic2019deep}.
GIN+EU baseline is similar to LaundroGraph~\cite{cardoso2022laundrograph}, which is a GNN specifically designed for anti-money laundering.
The accuracy results for these GNNs on the AML datasets are obtained from Altman et al.~\cite{altman_realistic_2023}.
Furthermore, all of the baselines, as well as our graph ML pipeline, are trained without the source and destination account IDs of the transactions.
This prevents the models from identifying money laundering transactions based on the memorisation of account IDs.

\textbf{Graph Feature Preprocessor setup.}
We configure GFP to extract the graph-based features in the following way.
The features are extracted from the AML datasets using a time window of six hours for scatter-gather patterns and a time window of one day for the rest of the graph-based features.
We specify a cycle-length constraint of $10$ for simple cycle enumeration. 
We use the "Amount" and "Timestamp" fields of the basic transaction features to generate the vertex-statistics-based features.
Feature extraction from the ETH Phishing dataset is performed using a $20$-day time window for all graph-based features.
In addition, we disable the generation of temporal cycles and specify a hop constraint of $5$ for simple cycle enumeration. We use the "Amount", "Timestamp", and "Block Nr." fields of the basic transaction features to generate the account statistics.
We selected these parameters after some careful exploration aimed at finding the best trade-offs between the throughput of GFP and the accuracy of the ML models used for scoring.

\begin{table}[t!]
\centering
\caption{Successive halving configurations used for hyperparameter tuning of both LightGBM and XGBoost models.}
\vspace{-.15in}
\addtolength{\tabcolsep}{-1.5pt}
\begin{tabular}{l|cccc}
\bottomrule
Datasets  & AML Small       & AML Medium      & AML Large  & ETH  \\ \hline
$x_0$  & $1000$        & $100$         & $16$ & $100$   \\
$\eta$ & $2$& $2$& $2$  & $2$   \\
$r_0$  & $0.1$        & $0.2$       & $0.2$ & $0.1$ \\
\toprule
\end{tabular}

\label{table:sh_config}
\vspace{-.1in}
\end{table}

\begin{table}[t!]
\centering
\caption{Model parameter ranges used at tuning time.}
\vspace{-.10in}
\addtolength{\tabcolsep}{-1.5pt}

\begin{tabular}{l|l|l|l}
    \bottomrule
    \multicolumn{2}{c|}{\textbf{LightGBM}} & \multicolumn{2}{c}{\textbf{XGBoost}}    \\ \hline
    \textbf{Parameter} & \textbf{Range} & \textbf{Parameter} & \textbf{Range} \\ \hline
    
    num\_round & $(10, 1000)$           & num\_round & $(10, 1000)$                  \\ \hline
    num\_leaves & $(1, 16384)$          & max\_depth & $(1, 15)$                      \\ \hline
    learning\_rate & $10^{(-2.5, -1)}$  & learning\_rate & $10^{(-2.5, -1)}$         \\ \hline
    lambda\_l2 & $10^{(-2, 2)}$         & lambda & $10^{(-2, 2)}$                    \\ \hline
    scale\_pos\_weight & $(1, 10)$       & scale\_pos\_weight & $(1, 10)$             \\ \hline
    lambda\_l1 & $10^{(0.01, 0.5)}$     & colsample\_bytree & $(0.5, 1.0)$            \\ \hline
    
     &                                  & subsample & $(0.5, 1.0)$                  \\ \hline
    \multicolumn{4}{c}{early\_stopping\_rounds = 20}                                 \\

    \toprule
\end{tabular}

\label{tab:hpt-ranges}
\vspace{-.15in}
\end{table}

\textbf{Graph ML pipeline training.}
The training step of our graph ML pipeline is illustrated in Figure~\ref{fig:graphml_pipeline}a.
First, the transactions available for training are ordered in ascending order of their timestamps and are split into train, validation, and test sets.
This split is performed in such a way that the transactions from the train set have the lowest timestamps and the transactions from the test set have the highest.
Then, the transactions from the train and validation sets are forwarded to GFP to generate the enriched graph-based features for the transactions from these two sets.
To prevent any form of information leakage at training time, the training set is processed before the validation set.
In that case, graph-based features for the transactions of the train set are computed on the graph created using only those transactions, and thus no information from the validation set is used.
Finally, the train and validation sets with enriched features are then used to train the gradient boosting  models~\cite{ke_lightgbm_2017,chen_xgboost_2016}. 

\textbf{Boosting machine parameter tuning.}
As part of training the gradient-boosting-based models, we perform hyper-parameter tuning using the successive halving approach~\cite{jamieson_best-arm_2014}.
This approach starts by randomly sampling $x_0$ model parameter combinations using a fraction $r_0 \leq 1$ of the train set.
Then, for a given $\eta$ > 1 parameter, the algorithm finds the best $x_0/\eta$ configurations, which are used in the next round of successive halving that uses $\eta \times r_0$ of the train set.
This process continues until the fraction of the training set used for evaluation reaches $1$.
The successive halving parameters used in our experiments are given in Table~\ref{table:sh_config} and the parameter ranges of LightGBM and XGBoost models used for hyperparameter tuning are given in Table~\ref{tab:hpt-ranges}.

\textbf{Graph ML inference.}
The inference step of our graph ML pipeline is shown in Figure~\ref{fig:graphml_pipeline}b.
First, we load the model trained using the setup shown in Figure~\ref{fig:graphml_pipeline}a.
Then, we initialise GFP by loading past financial transactions using the \textit{fit} function.
These past financial transactions are used to create the initial in-memory graph.
Next, the transactions from the test set are grouped into batches and forwarded to GFP using the \textit{transform} function.
This function updates the existing dynamic graph using the forwarded transactions and enriches those transactions with graph-based features of the same type as those generated in the train setup (see Figure~\ref{fig:graphml_pipeline}a).
Finally, the enriched test transactions are sent to the pre-trained machine learning model for detection of transactions associated with financial crime. 

\textbf{Data split.} To tune the parameters of the models and to test the model generalisation performance, we split the input data into train, validation, and test sets. The train and validation sets are used by the successive halving scheme to tune the model, while the test set is used for the final evaluation of the model. 
For AML datasets, the splitting is performed such that $60\%$ of transactions with the smallest timestamps is selected as a training set, the next $20\%$ transactions with the smallest timestamps excluding the ones from the training set are selected as a validation set, and the rest are selected as the test set.
For the ETH dataset, we define the timestamp of an account as the minimum timestamp among the transactions that involve this account and split the accounts of the dataset such that $65 \%$ of the accounts with the smallest timestamp exist only in the training set, the next $15 \%$ of the accounts exist only in the validation dataset, and the rest are in the test set.
Splitting the datasets in the aforementioned way prevents data leakage in our experiments.

\section{Results}
\label{sect:results}

\begin{table*}[th!]
\centering
\caption{Minority class F1 scores (\%) of the money laundering detection task using the AML datasets and the phishing detection task using the ETH Phishing dataset. NA stands for not available.
}
\vspace{-.10in}
\addtolength{\tabcolsep}{-1.5pt}
\resizebox{\linewidth}{!}{%
\begin{tabular}{l|c|ccc|ccc|c|c}
\bottomrule
\multirow{2}{*}{Model}  & \multirow{2}{*}{\Centerstack{batch\\size}} & \multicolumn{3}{c|}{AML HI} & \multicolumn{3}{c|}{AML LI}  & \multirow{2}{*}{\Centerstack{batch\\size}} & \multirow{2}{*}{ETH Phishing}   \\
  && Small& Medium& Large& Small& Medium& Large & & \\ \hline
 GIN~\cite{hu2019strategies}  &  $\infty$ & {\cellcolor[HTML]{8BCF89}} \color[HTML]{000000} 28.70 ± 1.13 & {\cellcolor[HTML]{3DA65A}} \color[HTML]{F1F1F1} 42.30 ± 0.44 &  NA & {\cellcolor[HTML]{E6F5E1}} \color[HTML]{000000} 7.90 ± 2.78 & {\cellcolor[HTML]{EFF9EB}} \color[HTML]{000000} 3.86 ± 3.62 &  NA &  $\infty$ & {\cellcolor[HTML]{95D391}} \color[HTML]{000000} 26.92 ± 7.52 \\
 GIN+EU~\cite{battaglia2018relational_edgeupdates, cardoso2022laundrograph} &  $\infty$ & {\cellcolor[HTML]{29914A}} \color[HTML]{F1F1F1} 47.73 ± 7.86 & {\cellcolor[HTML]{238B45}} \color[HTML]{F1F1F1} 49.26 ± 4.02 &  NA & {\cellcolor[HTML]{B4E1AD}} \color[HTML]{000000} 20.62 ± 2.41 & {\cellcolor[HTML]{E9F7E5}} \color[HTML]{000000} 6.19 ± 8.32 &  NA &  $\infty$ & {\cellcolor[HTML]{6DC072}} \color[HTML]{000000} 33.92 ± 7.34 \\
 PNA~\cite{corso_principal_2020} &  $\infty$ & {\cellcolor[HTML]{026F2E}} \color[HTML]{F1F1F1} 56.77 ± 2.41 & {\cellcolor[HTML]{006227}} \color[HTML]{F1F1F1} 59.71 ± 1.91 &  NA & {\cellcolor[HTML]{C7E9C0}} \color[HTML]{000000} 16.45 ± 1.46 & {\cellcolor[HTML]{90D18D}} \color[HTML]{000000} 27.73 ± 1.65 &  NA &  $\infty$ & {\cellcolor[HTML]{19833E}} \color[HTML]{F1F1F1} 51.49 ± 4.29 \\ \hline
 LightGBM~\cite{ke_lightgbm_2017} &  --- & {\cellcolor[HTML]{B1E0AB}} \color[HTML]{000000} 21.30 ± 0.30 & {\cellcolor[HTML]{BDE5B6}} \color[HTML]{000000} 18.60 ± 0.10 & {\cellcolor[HTML]{A2D99C}} \color[HTML]{000000} 24.50 ± 0.20 & {\cellcolor[HTML]{F3FAF0}} \color[HTML]{000000} 2.05 ± 0.81 & {\cellcolor[HTML]{F0F9ED}} \color[HTML]{000000} 3.3 ± 0.48 & {\cellcolor[HTML]{EFF9EB}} \color[HTML]{000000} 4.04 ± 0.16 &  --- & {\cellcolor[HTML]{D1EDCB}} \color[HTML]{000000} 13.74 ± 0.54 \\
 \textbf{GFP}+LightGBM&  128 & {\cellcolor[HTML]{005221}} \color[HTML]{F1F1F1} 62.86 ± 0.25 & {\cellcolor[HTML]{006328}} \color[HTML]{F1F1F1} 59.48 ± 0.15 & {\cellcolor[HTML]{00692A}} \color[HTML]{F1F1F1} 58.03 ± 0.19 & {\cellcolor[HTML]{B2E0AC}} \color[HTML]{000000} 20.83 ± 1.50 & {\cellcolor[HTML]{A0D99B}} \color[HTML]{000000} 24.74 ± 0.46 & {\cellcolor[HTML]{A5DB9F}} \color[HTML]{000000} 23.67 ± 0.11 &  128 & {\cellcolor[HTML]{46AE60}} \color[HTML]{F1F1F1} 40.17 ± 0.22 \\
 \textbf{GFP}+LightGBM &  2048 & {\cellcolor[HTML]{005E26}} \color[HTML]{F1F1F1} 60.52 ± 0.59 & {\cellcolor[HTML]{067230}} \color[HTML]{F1F1F1} 56.12 ± 0.37 & {\cellcolor[HTML]{0B7734}} \color[HTML]{F1F1F1} 54.76 ± 0.08 & {\cellcolor[HTML]{C0E6B9}} \color[HTML]{000000} 17.99 ± 0.60 & {\cellcolor[HTML]{B1E0AB}} \color[HTML]{000000} 21.06 ± 0.08 & {\cellcolor[HTML]{AADDA4}} \color[HTML]{000000} 22.65 ± 0.59 &  $\infty$ & {\cellcolor[HTML]{1C8540}} \color[HTML]{F1F1F1} 51.00 ± 1.01 \\ \hline
 XGBoost~\cite{chen_xgboost_2016} &  --- & {\cellcolor[HTML]{B8E3B2}} \color[HTML]{000000} 19.75 ± 0.89 & {\cellcolor[HTML]{B6E2AF}} \color[HTML]{000000} 20.10 ± 0.22 & {\cellcolor[HTML]{DCF2D7}} \color[HTML]{000000} 10.61 ± 6.73 &  0.21 ± 0.22 & {\cellcolor[HTML]{F6FCF4}} \color[HTML]{000000} 0.40 ± 0.14 &  0.00 ± 0.00 &  --- & {\cellcolor[HTML]{CBEAC4}} \color[HTML]{000000} 15.52 ± 0.15 \\
 \textbf{GFP}+XGBoost &  128 & {\cellcolor[HTML]{005020}} \color[HTML]{F1F1F1} 63.23 ± 0.17 & {\cellcolor[HTML]{00441B}} \color[HTML]{F1F1F1} 65.69 ± 0.26 & {\cellcolor[HTML]{3BA458}} \color[HTML]{F1F1F1} 42.68 ± 12.93 & {\cellcolor[HTML]{92D28F}} \color[HTML]{000000} 27.28 ± 0.69 & {\cellcolor[HTML]{7FC97F}} \color[HTML]{000000} 31.03 ± 0.22 & {\cellcolor[HTML]{A3DA9D}} \color[HTML]{000000} 24.23 ± 0.12 &  128 & {\cellcolor[HTML]{5AB769}} \color[HTML]{F1F1F1} 37.01 ± 2.45 \\
 \textbf{GFP}+XGBoost &  2048 & {\cellcolor[HTML]{00481D}} \color[HTML]{F1F1F1} 64.77 ± 0.47 & {\cellcolor[HTML]{006428}} \color[HTML]{F1F1F1} 59.19 ± 0.29 & {\cellcolor[HTML]{026F2E}} \color[HTML]{F1F1F1} 56.88 ± 0.21 & {\cellcolor[HTML]{8DD08A}} \color[HTML]{000000} 28.25 ± 0.80 & {\cellcolor[HTML]{B0DFAA}} \color[HTML]{000000} 21.36 ± 0.90 &  {\cellcolor[HTML]{AADDA4}} \color[HTML]{000000} 22.64 ± 0.15 &  $\infty$ & {\cellcolor[HTML]{228A44}} \color[HTML]{F1F1F1} 49.40 ± 0.54 \\
\toprule
\end{tabular}
}
\vspace{-.14in}
\label{tab:aml_accuracy}
\end{table*}

In this section, we evaluate the accuracy of our graph ML pipeline and other baselines trained on the datasets from Table~\ref{tab:datasets}.
We refer to our graph ML pipeline that uses LightGBM and XGBoost as GFP+LightGBM and GFP+XGBoost, respectively.
As a measure of accuracy, we use the minority-class F1 score.
The F1 scores reported are averaged across five different runs. 
The standard deviation of the F1 score is also reported for each experiment.

Our graph ML pipeline requires transactions to arrive in batches.
For the AML datasets, the graph ML pipeline uses batch sizes of $128$ and $2048$.
In addition, for the ETH Phishing dataset, graph feature extraction is performed using batch sizes of $128$ and $\infty$. 
When using a batch size of $\infty$, all the transactions of the test set are made available to GFP in a single batch. 
Using a batch size of $\infty$ essentially corresponds to an offline solution and, in principle, can lead to better accuracy because, in this case, the future transactions are also visible during feature extraction. 
However, if real-time processing capability is required by an application, the batch size will have to be constrained.
Note that GNN baselines require the entire dataset to be available in memory, making it effectively an offline solution with batch size $\infty$.

\textbf{AML results.} The minority class F1 scores of the ML models that perform laundering detection using  AML datasets are shown in Table~\ref{tab:aml_accuracy}.
Clearly, our graph-based features lead to significant improvements in the F1 scores achieved by gradient-boosting models.
Without our graph-based features, the maximum F1 score that LightGBM and XGBoost achieve is $24.5\%$ for the AML HI datasets and $4.04\%$ for the AML LI datasets.
The reason for this low accuracy is that the labels in AML datasets are highly imbalanced, and the number of illicit transactions in these datasets is at most $0.13\%$ of the total number of transactions (see Table~\ref{tab:datasets}).
Our graph ML pipeline, in which LightGBM and XGBoost models use our graph-based features in addition to basic features, achieves up to a $46\%$ higher F1 scores than the models that use only basic features.
Furthermore, our graph ML pipeline that uses XGBoost models consistently achieves higher F1 scores than GNN baselines.
Compared to PNA, the GNN baseline with the highest accuracy, our graph ML pipeline with XGBoost achieves up to an $8\%$ higher F1 score for AML HI datasets and up to an $11.8\%$ higher F1 score for LI datasets.

 \begin{figure}[t!]
    \centering
    \includegraphics[width=1.0\columnwidth]{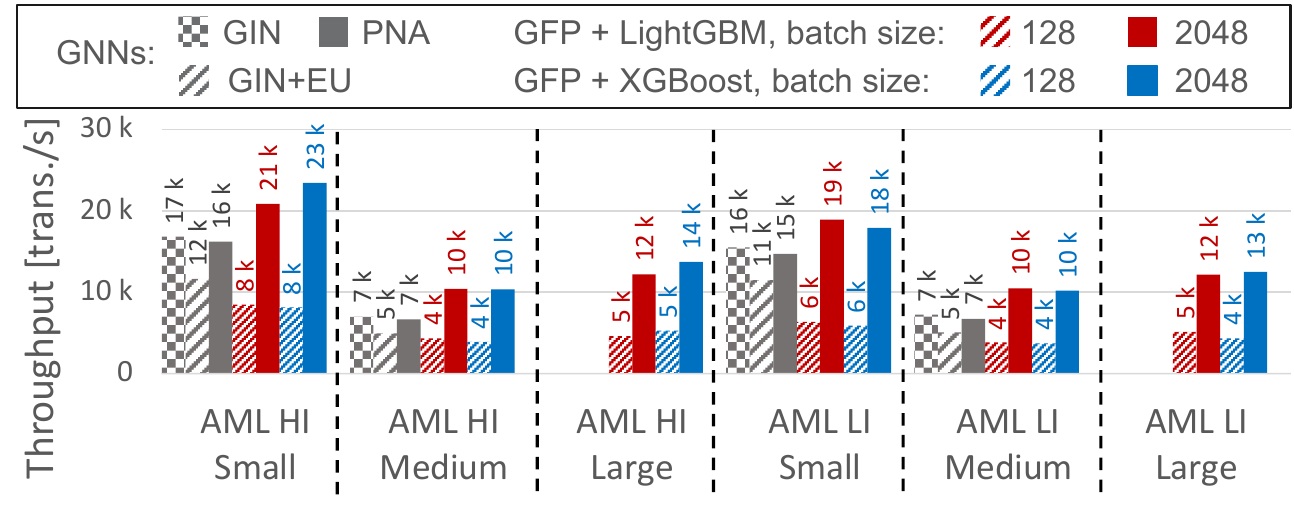}
\vspace{-.2in}
    \caption{Our graph ML pipeline has higher throughput compared to GNN baselines executed on a V100 GPU.}
    \label{fig:aml_performance}
\vspace{-.2in}
\end{figure}

 \begin{figure}[t]
    \centering
	\centerline{
        \includegraphics[width=1.0\columnwidth]{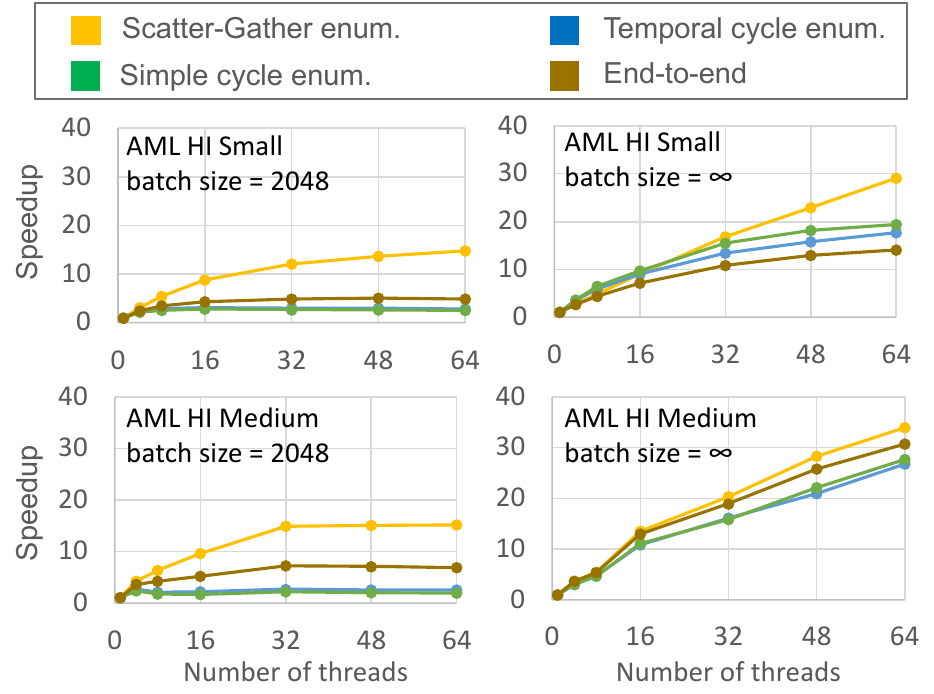}
    }
    \vspace{-.1in}
    \caption{Scalability of executing different parts of our GFP library, as well as its end-to-end execution. The speedup is relative to the single-threaded execution.}
    \vspace{-.25in}
    \label{fig:scalability}
\end{figure}

The effect of different types of graph-based features produced by GFP on the accuracy of our graph ML pipeline for the AML task is shown in Table~\ref{tab:ablation}. 
We observe that including graph features based on fan-in and fan-out patterns already improves the minority class F1 score by more than $30\%$ compared to the case that uses only basic transaction features.
Including multi-hop graph pattern features, i.e., features based on cycles and scatter-gather patterns, further improves the F1 score by up to $4\%$.
Finally, by incorporating vertex-statistics-based features produced by GFP, our graph ML pipeline is able to achieve higher accuracy compared to the PNA baseline (see Table~\ref{tab:aml_accuracy}).
Thus, each type of graph-based feature contributes to the overall accuracy of our graph ML pipeline.

\begin{table*}[t]
\centering
\caption{Minority class F1 scores (\%) of our graph ML pipeline demonstrating the effect of different graph-based features produced by GFP on the accuracy of money laundering detection. Multi-hop pattern features include features based on simple cycles, temporal cycles, and scatter-gather patterns.
}
\vspace{-.10in}
\addtolength{\tabcolsep}{-1.5pt}
\resizebox{\linewidth}{!}{%
\begin{tabular}{l|cc|cc|cc|cc|c|c}
\bottomrule
Dataset                           & \multicolumn{4}{c|}{AML HI Small}                           & \multicolumn{4}{c|}{AML HI Medium}    & \multicolumn{2}{c}{ETH Phishing}                       \\ \hline
Model                             & \multicolumn{2}{c|}{LightGBM} & \multicolumn{2}{c|}{XGBoost} & \multicolumn{2}{c|}{LightGBM} & \multicolumn{2}{c|}{XGBoost} & \multicolumn{1}{c|}{LightGBM} & \multicolumn{1}{c}{XGBoost} \\ \hline
batch size                        & 128           & 2048         & 128           & 2048         & 128           & 2048         & 128          & 2048  & $\infty$ & $\infty$        \\ \hline
basic features  & {\cellcolor[HTML]{B1E0AB}} \color[HTML]{000000} 21.30 ± 0.30 & {\cellcolor[HTML]{B1E0AB}} \color[HTML]{000000} 21.30 ± 0.30 & {\cellcolor[HTML]{B8E3B2}} \color[HTML]{000000} 19.75 ± 0.89 & {\cellcolor[HTML]{B8E3B2}} \color[HTML]{000000} 19.75 ± 0.89 & {\cellcolor[HTML]{BDE5B6}} \color[HTML]{000000} 18.60 ± 0.10 & {\cellcolor[HTML]{BDE5B6}} \color[HTML]{000000} 18.60 ± 0.10 & {\cellcolor[HTML]{B6E2AF}} \color[HTML]{000000} 20.10 ± 0.22 & {\cellcolor[HTML]{B6E2AF}} \color[HTML]{000000} 20.10 ± 0.22 & {\cellcolor[HTML]{D1EDCB}} \color[HTML]{000000} 13.74 ± 0.54 & {\cellcolor[HTML]{CBEAC4}} \color[HTML]{000000} 15.52 ± 0.15 \\
+ fan-in/fan-out features & {\cellcolor[HTML]{1C8540}} \color[HTML]{F1F1F1} 50.85 ± 0.83 & {\cellcolor[HTML]{218944}} \color[HTML]{F1F1F1} 49.73 ± 1.20 & {\cellcolor[HTML]{026F2E}} \color[HTML]{F1F1F1} 56.88 ± 0.66 & {\cellcolor[HTML]{006227}} \color[HTML]{F1F1F1} 59.71 ± 0.07 & {\cellcolor[HTML]{2C944C}} \color[HTML]{F1F1F1} 46.71 ± 0.17 & {\cellcolor[HTML]{1D8640}} \color[HTML]{F1F1F1} 50.59 ± 0.36 & {\cellcolor[HTML]{137D39}} \color[HTML]{F1F1F1} 53.00 ± 0.08 & {\cellcolor[HTML]{097532}} \color[HTML]{F1F1F1} 55.25 ± 0.19 & {\cellcolor[HTML]{62BB6D}} \color[HTML]{F1F1F1} 35.92 ± 1.96 & {\cellcolor[HTML]{45AD5F}} \color[HTML]{F1F1F1} 40.46 ± 0.94\\
+ multi-hop pattern features & {\cellcolor[HTML]{0C7735}} \color[HTML]{F1F1F1} 54.66 ± 0.39 & {\cellcolor[HTML]{087432}} \color[HTML]{F1F1F1} 55.54 ± 0.55 & {\cellcolor[HTML]{006729}} \color[HTML]{F1F1F1} 58.60 ± 0.15 & {\cellcolor[HTML]{005B25}} \color[HTML]{F1F1F1} 61.01 ± 0.24 & {\cellcolor[HTML]{2A924A}} \color[HTML]{F1F1F1} 47.47 ± 0.21 & {\cellcolor[HTML]{19833E}} \color[HTML]{F1F1F1} 51.40 ± 0.15 & {\cellcolor[HTML]{097532}} \color[HTML]{F1F1F1} 55.42 ± 0.23 & {\cellcolor[HTML]{077331}} \color[HTML]{F1F1F1} 55.92 ± 0.26 &
{\cellcolor[HTML]{4BB062}} \color[HTML]{F1F1F1} 39.46 ± 0.27 &{\cellcolor[HTML]{3BA458}} \color[HTML]{F1F1F1} 42.76 ± 0.48\\
+ vertex-statistic-based features & {\cellcolor[HTML]{005221}} \color[HTML]{F1F1F1} 62.86 ± 0.25 & {\cellcolor[HTML]{005E26}} \color[HTML]{F1F1F1} 60.52 ± 0.59 & {\cellcolor[HTML]{005020}} \color[HTML]{F1F1F1} 63.23 ± 0.17 & {\cellcolor[HTML]{00481D}} \color[HTML]{F1F1F1} 64.77 ± 0.47 & {\cellcolor[HTML]{006328}} \color[HTML]{F1F1F1} 59.48 ± 0.15 & {\cellcolor[HTML]{067230}} \color[HTML]{F1F1F1} 56.12 ± 0.37 & {\cellcolor[HTML]{00441B}} \color[HTML]{F1F1F1} 65.70 ± 0.26 & {\cellcolor[HTML]{006428}} \color[HTML]{F1F1F1} 59.19 ± 0.29 & {\cellcolor[HTML]{1C8540}} \color[HTML]{F1F1F1} 51.00 ± 1.01 & {\cellcolor[HTML]{228A44}} \color[HTML]{F1F1F1} 49.40 ± 0.54  \\
\toprule
\end{tabular}
}
\vspace{-.15in}
\label{tab:ablation}
\end{table*}

Figure~\ref{fig:aml_performance} shows the throughput of our graph ML pipeline and GNN baselines.
The performance of our graph ML pipeline is evaluated using $64$ software threads of the Cascade Lake Intel Xeon Processor available from IBM Cloud~\cite{ibm_cloud}, and the performance of GNN baselines is evaluated on an NVIDIA Tesla V100 GPU.
We observe that our graph ML pipeline is able to achieve higher throughput than GNN baselines when it receives transactions in batches of $2048$.
This throughput is the result of the scalable parallel graph pattern mining algorithms that GFP uses, as shown in Figure~\ref{fig:scalability}.
This figure also shows that our streaming scatter-gather algorithm, introduced in Section~\ref{sect:gpm}, scales almost linearly with the number of software threads when batch size is infinity.
As a result of this scalability, the average latency of processing batches of 128 and 2048 transactions from the AML dataset is $30$ ms and $148$ ms, respectively.
Being able to process a batch of transactions with low latency makes GFP suitable for real-time processing.

 \begin{figure}[t!]
    \centering
    \includegraphics[width=1.0\columnwidth]{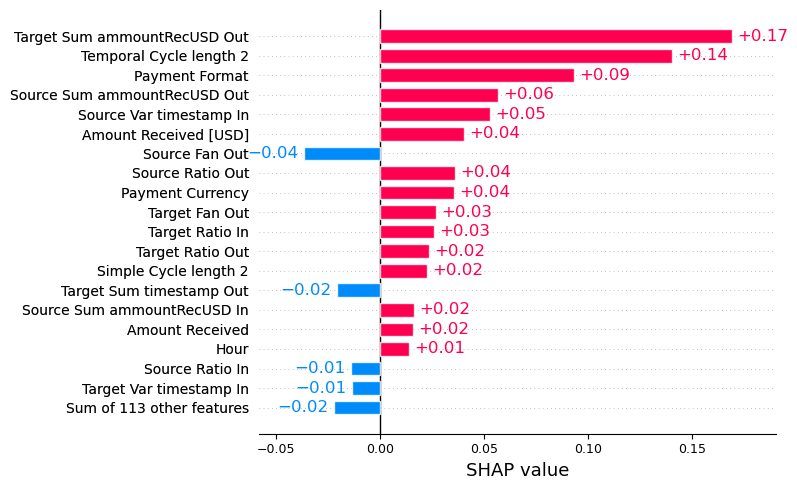}
\vspace{-.25in}
    \caption{Importance of features used by our GFP+LightGBM setup to flag an AML HI Small transaction as illicit.}
    \label{fig:explainability}
\vspace{-.2in}
\end{figure}

\textbf{Explainability.}  A benefit of our graph ML pipeline is that it produces explainable results. Using the SHAP library~\cite{shap}, we can obtain the importance of features a gradient-boosting-based model used to flag a transaction as illicit.
For example, the two most important features used for flagging a transaction as illicit in Figure~\ref{fig:explainability} are the number of two-hop temporal cycles (\textit{Temporal Cycle length~2}) and a vertex statistics feature, which represents the sum of money the target account~received (\textit{Target Sum ammountRecUSD~Out}).
Explaining the decision is critical for increasing trust in a fraud detection system because it allows analysts to verify the decisions of the system as needed.

\textbf{ETH Phishing  results.} Table~\ref{tab:aml_accuracy} also shows the minority class F1 scores achieved by the ML models we trained on the ETH Phishing dataset to perform phishing detection. 
When using a batch size of $128$, our graph-based features enable F1-score improvements exceeding $20 \%$ for both  LightGBM and XGBoost.
Setting the batch size to $\infty$ further improves the F1 score of LightGBM  to $51\%$. 
In that case, LightGBM with our graph-based features outperforms the GIN+EU baseline by $10\%$ and achieves competitive accuracy with PNA.
However, increasing the batch size from $128$ to $\infty$ effectively makes our graph ML pipeline an offline solution.
In general, the optimal configuration of GFP depends on the requirements of the end application and might require trading off performance~for~accuracy.

\section{Related Work}

\textbf{Graph machine learning} has applications in many different fields, including financial transaction network analysis~\cite{NichollsIEEEAccess2021,liu_knowledge_2021,ChangIEEETSMCS2020,WangArxiv2021}, fraud detection~\cite{LiuWWW2021, zhu_modeling_2020, cao_titant_2019, eddin_anti-money_2022, aws_fraud, forrester_wave}, drug discovery~\cite{GaudeletBriefings2021}, molecular property prediction~\cite{ZhangArxiv2021}, genomics~\cite{Schulte-SasseNatureMI2021}, recommender systems~\cite{EksombatchaiWWW2018}, social network analysis~\cite{BensonScience2016,FanWWW2019}, and relation prediction in knowledge graphs~\cite{QinAAAI2021}.
Fraud detection systems TitAnt~\cite{cao_titant_2019} and Eddin et al.~\cite{eddin_anti-money_2022} are graph machine learning systems that extract features from transaction graphs by generating node embeddings~\cite{perozzi_deepwalk_2014} or by performing random walks~\cite{oliveira_guiltywalker_2021} in graphs.
These features are then used by machine learning models to predict whether an incoming transaction is fraudulent or~not.

\textbf{Graph neural networks} (GNNs)~\cite{xu2018powerful, velickovic2018graph, bouritsas_improving_2023, kipf2017semisupervised, hamilton2017inductive,cardoso2022laundrograph, LiuWWW2021, WangArxiv2021} are powerful tools that can be used for the purpose of financial crime detection.
Cardoso et al.~\cite{cardoso2022laundrograph} and Weber et al.~\cite{weber2019anti} apply GNN to the anti-money laundering problem, Kanezashi et al.~\cite{kanezashi_ethereum_2022} apply GNN to the phishing detection problem on the Ethereum blockchain, and Rao et al.~\cite{rao_xfraud_2021} uses a GNN to detect fraudulent transactions.
Graph Substructure Network, proposed by Bouritsas et al.~\cite{bouritsas_improving_2023}, takes advantage of pre-calculated subgraph pattern counts to improve the expressivity of GNNs.
GNNs could also be used to count subgraph patterns, such as in Chen et al. ~\cite{chen_substructures_2020}, which could enable detecting patterns associated with financial crime.
In contrast to our work, GNNs cannot straightforwardly operate in a streaming manner and require the entire dataset to be available at the time of~testing.

\textbf{Dynamic graph management} is often required for real-time processing of financial transactions.
Dynamic graph data structures, such as STINGER~\cite{ediger_stinger_2012}, GraphTinker~\cite{jaiyeoba_graphtinker_2019}, and Sortledton~\cite{fuchs_sortledton_2022} enable dynamic insertions of edges into the graph as well as their removal from the graph.
However, STINGER and GraphTinker cannot be directly used for representing financial transaction graphs because they do not support the maintenance of multiple edges with the same source and destination vertices.
In-memory graph databases~\cite{zhu_livegraph_2020, buragohain_a1_2020, carter_nanosecond_2019} can also be used for dynamic graph management.
Bing's distributed in-memory graph database A1~\cite{buragohain_a1_2020} leverages high-speed Remote Direct Memory Access to maintain an evolving graph containing billions of vertices and edges.
Linkedin's in-memory graph database~\cite{carter_nanosecond_2019} enables low latency read and write operations to the graph and supports the representation of N-ary relationships in the graphs.
Our dynamic graph data structure does not require support for N-ary relationships, and thus can be implemented in a simpler manner.

\section{Conclusions}

We presented Graph Feature Preprocessor (GFP), a software library for fast feature extraction from dynamically changing transaction graphs.
To achieve fast feature extraction, our library leverages an in-memory dynamic multigraph representation as well as fine-grained parallel subgraph enumeration algorithms.
GFP enables our graph ML pipeline to operate in a streaming manner with low per-batch latency and higher throughput compared to the GNN baselines presented in the experiments.
This capability makes GFP suitable for scenarios that require real-time processing.

We have also shown that the graph-based features generated by GFP can significantly improve the accuracy of gradient-boosting-based machine learning models.
The graph-based features improve the minority class F1 score of gradient-boosting-based machine learning models by up to $46\%$ for the synthetic AML datasets and by up to $35\%$ for a real-world phishing detection dataset extracted from Ethereum.
Furthermore, we show that our solution achieves up to a $36\%$ higher F1 score than GNN baselines for the AML task.
In particular, our graph ML pipeline achieves up to a $24\%$ higher minority-class F1 score compared to the GIN+EU baseline with the similar architecture to LaundroGraph~\cite{cardoso2022laundrograph}, which is a GNN designed specifically for anti-money laundering. 

The application scope of our GFP library is not limited to money laundering detection. 
Given that a cycle in a graph can be an indicator of tax avoidance~\cite{hajdu_temporal_2020}, circular trading~\cite{palshikar_collusion_2008, islam_approach_2009, jiang_trading_2013}, and credit card frauds~\cite{qiu_real-time_2018, nicholls_financial_2021}, a GFP could also help to detect these types of frauds.
However, the reliance on pre-defined subgraph patterns, such as cycles, is one drawback of this library, which we plan to address as part of the future work by adding the support for subgraph matching using user-defined subgraph patterns in GFP~\cite{sun_in-memory_2020}.
Furthermore, we plan to add support for feature extraction based on additional subgraph patterns, such as cliques~\cite{bron_algorithm_1973} and bicliques~\cite{prisner_bicliques_2000}.
Being able to enumerate these patterns could enable the detection of close-knit communities~\cite{lu_community_2018} as well as stacked money laundering patterns~\cite{altman_realistic_2023} encountered in various different financial crime~scenarios.

\begin{acks}
The support of Swiss National Science Foundation (project number 172610) for this work is gratefully acknowledged. The authors would like to thank Donna Eng Dillenberger, Thomas Parnell, Martin Petermann, Evan Rivera, and Elpida Tzortzatos from IBM for their support, feedback, and suggestions during the course of this~work.
\end{acks}

\balance
\bibliographystyle{ACM-Reference-Format}
\bibliography{References}

\end{document}